\documentclass[runningheads]{llncs}

\usepackage[T1]{fontenc}
\usepackage{graphicx}
\usepackage{booktabs}
\usepackage{microtype}   
\usepackage{hyperref}

\graphicspath{{figures/}{./}}

\begin{document}

\title{AI-Driven Collaborative Assembly Line Inspection:
System Integration and Deployment Challenges}
\titlerunning{AI-Driven Collaborative Assembly Line Inspection}

\author{Asya \"Unal \and
Amr Okasha\thanks{Corresponding author: okasha@teknopar.com.tr} \and
Ege \c{C}\i rakman \and
Perin \"Unal}
\authorrunning{A. \"Unal et al.}

\institute{Department of Research and Development, TEKNOPAR, Ankara, T\"urkiye\\
\email{aunal@teknopar.com.tr} \\
\email{okasha@teknopar.com.tr} \\ 
\email{cirakman18@itu.edu.tr} \\ 
\email{punal@teknopar.com.tr}}

\maketitle

\begin{abstract}
Manual visual inspection on assembly lines is a persistent manufacturing
bottleneck: operator fatigue over extended shifts lowers defect-detection
rates. This paper presents the design, integration, and field deployment of
an AI-assisted collaborative inspection cell at the Silverline
kitchen-appliance factory, developed within the AI-PRISM project. The cell
couples a Universal Robots UR\,10e cobot carrying a machine-vision
defect-detection pipeline with a Comau Racer-5 cobot for functional tests,
coordinated through ROS\,2 Humble on an Ubuntu~22.04\,LTS server.
Multi-modal data---Basler camera imagery, TIA microphone acoustics, and SPS
electrical-safety measurements---are logged locally and visualised in real
time with Grafana. We report the practical deployment challenges---%
close-proximity safety, AI robustness under glare and reflections, ROS\,2
namespace collisions across two cobots, and operating-system and dependency
issues---together with the engineering solutions adopted, and structure the
integration through a four-level Human--Robot Interaction analysis. The
deployed cell cuts per-unit quality-check time from 82\,s to 61\,s
($\approx$25\%), raises final-control resource efficiency from 0.75 to 0.88,
reduces operator visual-inspection viewing time by 82\%, and significantly
lowers operator mental demand ($p=0.005$, NASA-TLX).

\keywords{Collaborative Robots \and Defect Detection \and
Human--Robot Interaction \and Robot Operating System \and Industry 5.0}
\end{abstract}

\section{Introduction}
\label{sec:intro}

Under Industry~4.0, quality assurance on high-throughput assembly lines still
relies heavily on human operators who visually inspect surfaces under fixed
lighting. Over extended shifts, sustained visual attention causes operator
fatigue that degrades detection accuracy and introduces
inconsistency~\cite{ref_fatigue}, motivating active research into augmenting
this task with computer vision and collaborative robotics.

Collaborative robots (cobots) share workspace with operators without full
safety enclosures, making them attractive for retrofitting existing
lines~\cite{ref_cobot_survey}. Paired with AI vision modules, a cobot can
perform camera-guided inspection autonomously while the operator retains tasks
requiring dexterity or judgement. Such a configuration does not remove the
operator but reallocates cognitive load from repetitive scanning toward
higher-value activity, following a human-centred design
philosophy~\cite{ref_human_centered}. However, moving integrated AI--cobot
systems from the laboratory to the factory floor raises challenges that are
underrepresented in the literature: most work reports algorithm performance in
controlled settings, while deployment-stage issues---driver incompatibilities,
sensor synchronisation, lighting variability, and safety-compliant motion
planning---receive little attention.

This paper reports on the \emph{Silverline pilot}, an industrial deployment
carried out as part of the AI-PRISM project~\cite{ref_aiprism_ra}, integrating
a UR\,10e cobot with an AI-based defect-detection pipeline on a
kitchen-appliance assembly line at Silverline Company in Turkey. The principal
contributions are threefold:
\begin{enumerate}
    \item an end-to-end system architecture for AI-assisted collaborative
          inspection, encompassing sensing, actuation, data acquisition, and
          real-time visualization;
    \item a structured account of the field-deployment challenges encountered
          and the engineering solutions adopted, organized by category
          (safety, AI robustness, and infrastructure);
    \item a four-level HRI analysis that maps the integration from individual
          task design to overall process coordination.
\end{enumerate}

The remainder of the paper surveys related work
(Section~\ref{sec:related}), details the system architecture
(Section~\ref{sec:architecture}), reports the deployment challenges and
solutions (Section~\ref{sec:challenges}), presents the experimental results
(Section~\ref{sec:experiments}), and concludes with future work
(Section~\ref{sec:conclusion}).

\section{Related Work}
\label{sec:related}

\subsubsection{Cobots in manufacturing inspection.}
The UR and COMAU collaborative ranges are widely used in quality control
because built-in power- and force-limiting enable fenceless operation alongside
operators. Collaboration is governed by ISO~10218-1~\cite{ref_iso10218} and
ISO/TS~15066~\cite{ref_iso15066}, the latter defining collaborative modes and
biomechanical contact thresholds. P\'erez \emph{et~al.}~\cite{ref_perez_aero}
report productivity and ergonomic gains from symbiotic HRC cells in aerospace
manufacturing, and surveys confirm that inspection and end-of-line testing are
common cobot applications whose success hinges on integrating perception and
safety subsystems~\cite{ref_cobot_survey}.

\subsubsection{AI-based visual defect detection.}
Convolutional neural networks have largely displaced hand-crafted features for
surface-defect inspection. Single-stage YOLO detectors~\cite{ref_yolo} suit
inline inspection by predicting bounding boxes and class scores in one forward
pass. Surveys report precision and recall above 90\% on diverse substrates
given representative training data, with transfer learning and augmentation
compensating for the small datasets typical of industry~\cite{ref_defect_survey};
specular reflections, non-uniform illumination, and rare defect classes remain
open problems.

\subsubsection{ROS and ROS\,2 in manufacturing.}
ROS is a \emph{de facto} integration substrate for research robotics; ROS\,2~%
\cite{ref_ros2} adds DDS-based middleware, improved real-time behaviour, and
security features suited to industrial use. The ROS-Industrial
consortium~\cite{ref_ros_industrial} has matured manipulator drivers,
calibration tooling, and motion-planning packages, and MoveIt!\,2~%
\cite{ref_moveit2} provides collision-aware trajectory generation for UR-series
and other collaborative arms.

\subsubsection{HRI frameworks for Industry~4.0/5.0.}
Rodr\'iguez-Guerra \emph{et~al.}~\cite{ref_hri_levels} propose a four-level HRI
taxonomy---task design, operation, work cell, and process---that separates the
design choices made at each abstraction level; AI-PRISM adopts it to structure
architectural and validation activities~\cite{ref_aiprism_ra}. Lario
\emph{et~al.}~\cite{ref_lario_social} analyse five AI-PRISM pilots and show that
well-designed HRC cells reduce musculoskeletal load and improve psychological
comfort, consistent with Industry~5.0's emphasis on worker well-being.

\subsubsection{Gap.}
Despite the maturity of the individual building blocks, few publications
systematically report the field-deployment challenges of combined AI+cobot
systems on real production lines~\cite{ref_rueda_barriers}: driver and OS
compatibility, dependency resolution, in-situ lighting calibration, and the
synchronisation of heterogeneous sensor streams are usually confined to internal
reports. This paper helps close that gap by documenting a multi-modal inspection
cell on a live kitchen-appliance line.

\section{System Architecture and Design}
\label{sec:architecture}

This section describes the hardware and software components of the Silverline
inspection cell and the interaction design that governs human--robot
collaboration within it.

\subsection{System Overview}
\label{sec:arch_overview}

The inspection cell is built around a Universal Robots UR\,10e collaborative
manipulator carrying an industrial Basler camera and a dedicated lighting
fixture on its end effector, operating alongside a human operator on a
kitchen-appliance assembly line. An SPS power test device performs
electrical-safety tests (grounding and functional checks) on each product, and
a microphone sensor captures acoustic data for anomaly detection. All devices
communicate over a local Ethernet network and are coordinated by a central
server running Ubuntu~22.04\,LTS with ROS\,2 Humble.

\begin{figure}[t]
\centering
\includegraphics[width=\textwidth]{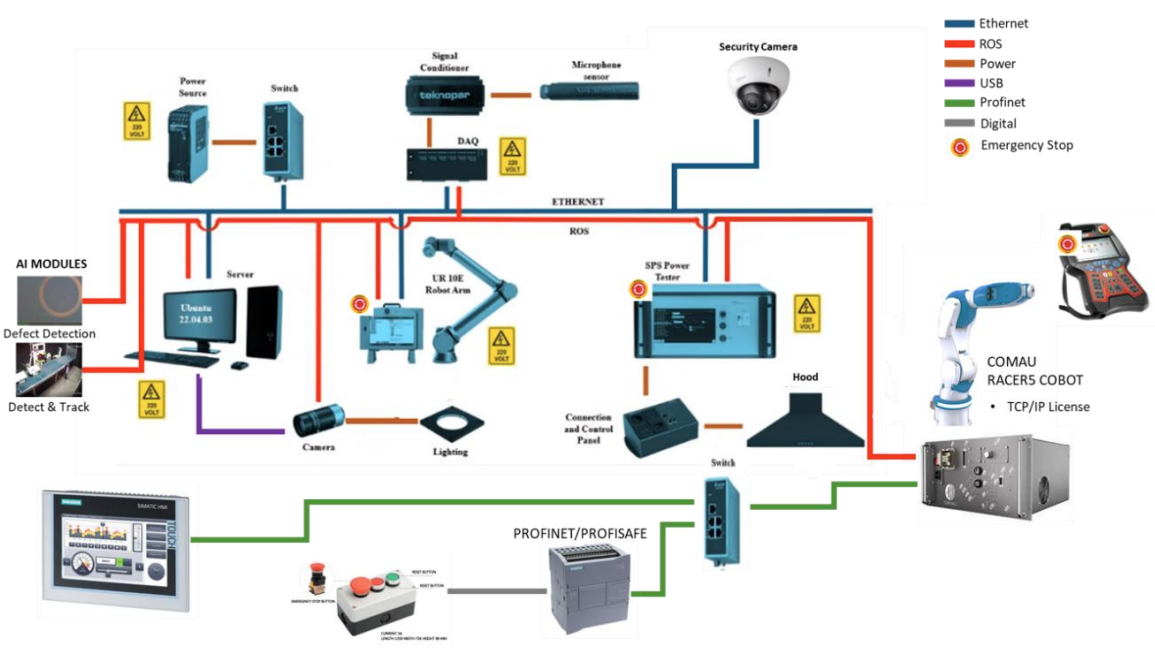}
\caption{Complete system architecture of the Silverline inspection cell. Blue
lines: Ethernet; red: ROS\,2 communication; orange: power supply; purple: USB;
green: Profinet/Profisafe; grey: digital I/O. The functional-safety subsystem
(bottom) comprises a Siemens S7-1200 PLC, emergency-stop buttons, and an HMI
panel. AI modules for defect detection and object tracking are hosted on the
central server.}
\label{fig:architecture}
\end{figure}

Fig.~\ref{fig:architecture} shows the complete system architecture. The server,
UR\,10e robot arm, DAQ module, and SPS test device are interconnected through
industrial Ethernet switches. A Teknopar signal conditioner interfaces the
microphone sensor with the DAQ. ROS\,2 topics carry control commands and sensor
data between nodes. A Profinet/Profisafe layer, managed by a Siemens S7-1200 PLC
with an HMI panel, implements functional safety including emergency-stop logic
and safe-speed monitoring. The architecture also accommodates a COMAU RACER5
cobot for basic functional tests (connected via TCP/IP) and a security camera
for supervisory monitoring and product tracking. AI modules for defect detection
and object tracking run on the server and consume the camera feed in real time.

\subsection{Data Acquisition and Processing}
\label{sec:arch_daq}

Data acquisition is handled by device-specific software packages running on the
central server. Audio data from the microphone sensor is captured using the
LabJack packages providing onset detection and spectral analysis capabilities.
Serial communication with the SPS power test device is managed through pySerial,
which reads grounding resistance and measures the power consumption of different
functions. Image frames from the Basler industrial camera are acquired via the
Basler pylon SDK, which provides hardware-triggered capture and exposure
control.

All acquired data are recorded in a local database with timestamped entries,
enabling post-hoc analysis and traceability. For real-time monitoring, the data
stream is forwarded to an InfluxDB time-series database, and Grafana dashboards
present the information as tables, charts, and trend plots, allowing operators
and engineers to observe the state of the system at a glance and to identify
anomalies during production.

\subsection{Robot Control and Motion Planning}
\label{sec:arch_robot}

Robot control is managed through the ROS\,2 framework. The MoveIt!\,2
motion-planning library provides collision-aware trajectory generation, enabling
the UR\,10e to navigate around workpieces and fixtures without manual waypoint
programming for every new product variant. For external control, URCaps allows
external modules to communicate the generated trajectories, which are then used
to control the robot's motion. Additionally, a remote-control package within the
ROS\,2 ecosystem allows precise teleoperation of the robot from external
computing devices when manual intervention is required.

Prior to autonomous operation, grounding and functional tests are manually
conducted using the SPS test device. These tests validate the electrical safety
and mechanical integrity of the robot system, and their results are logged
alongside production data in the central database.

\subsection{AI-Based Defect Detection}
\label{sec:arch_ai}

The perception stack comprises two AI modules. A hood-tracking module runs on
the supervisory security-camera feed: a YOLOv8 object detector~\cite{ref_yolo},
trained on a curated dataset of black and white hood images captured from that
viewpoint, detects every hood that enters the work cell and persists its
bounding box across frames so that each hood receives a stable identifier as it
moves along the conveyor. An active region of interest in front of the UR\,10e
triggers the inspection sequence when a hood remains stationary inside it for
more than three seconds; the tracker also exposes a counter of finished products
to the supervisory dashboard.

The second module---2D defect detection---is a YOLOv8-based detector that
processes high-resolution frames from the Basler camera on the UR\,10e end
effector. As the robot traverses a programmed inspection path over the glass and
metal surfaces of the hood, frames are streamed to the server, and the detector
classifies surface regions as defective or acceptable, recording the pixel
coordinates of every flagged region in the local database. The model was trained
on a dataset of representative hoods scanned under the final illumination
configuration of the cell (see Section~\ref{sec:chall_ai}), and its output is
overlaid on the operator user interface together with an audible alert when a
defect is found.

A central design objective of this subsystem is to reduce operator eye fatigue.
In conventional inspection, operators must visually scan every surface of every
product; the AI module automates this scanning step and flags only those regions
that require human judgment. This preserves the operator's role in the
process---the system augments rather than replaces the human worker---while
reallocating cognitive effort toward decision-making on flagged defects and
other quality-critical assessments.

\subsection{Human--Robot Interaction Design}
\label{sec:arch_hri}

The integration of ROS\,2 at Silverline can be analyzed through the four-level
HRI framework of Rodr\'iguez-Guerra \emph{et~al.}~\cite{ref_hri_levels}:

\paragraph{Task Design Level.}
Robots operate with varying degrees of autonomy while maintaining safe
collaboration with operators; safety protocols and interaction mechanisms were
evaluated so that inspection tasks run both efficiently and securely. Task
allocation is human-centered: the cobot handles repetitive scanning while the
operator retains authority over accept/reject decisions.

\paragraph{Operation Level.}
ROS\,2 orchestrates the robot's actions during each inspection cycle---%
navigation within the work cell, obstacle avoidance via MoveIt!\,2 planning, and
sequencing between the operator's manual steps and the robot's automated scanning
passes.

\paragraph{Work Cell Level.}
ROS\,2 topic scheduling plans and sequences the inspection, testing, and
data-logging subtasks to minimise idle time and reduce the risk of human--robot
spatial conflicts.

\paragraph{Process Level.}
At the highest level, ROS\,2 aligns the cell's throughput with Silverline's
overall production targets, synchronising robotic and human activities so that
the inspection cell does not become a bottleneck in the broader assembly line.

\section{Deployment Challenges and Solutions}
\label{sec:challenges}

Field deployment of the Silverline inspection cell surfaced a range of
challenges spanning safety, AI robustness, and infrastructure. This section
categorizes these challenges and describes the solutions adopted.

\subsection{Safety and Risk Mitigation}
\label{sec:chall_safety}

Three categories of safety risk were identified during deployment.

\paragraph{Close-proximity risks.}
The constrained workspace at Silverline requires the UR\,10e and human operators
to work in close proximity, where physical collision is the primary hazard.
Mitigation measures include enforced speed limits on robot motions (compliant
with ISO/TS\,15066 collaborative speed thresholds), clear demarcation of human
and robot work zones within the cell, and mandatory operator training on safe
interaction procedures~\cite{ref_iso15066}.

\paragraph{AI-driven behavior risks.}
Robots relying on AI for decision-making can exhibit unexpected actions when the
model encounters out-of-distribution inputs, particularly in dynamic or poorly
lit environments. To mitigate this risk, the AI model's training set was expanded
with diverse samples, and a continuous monitoring loop was implemented to flag
low-confidence predictions before they are translated into robot commands.

\paragraph{Collision prevention.}
Beyond speed limiting, the MoveIt!\,2 planning pipeline performs real-time
collision checking against a model of the work cell geometry. When an obstacle is
detected in the planned path, the planner generates an alternative trajectory or
reduces the robot's speed. The emergency-stop circuit, managed by the Siemens PLC
via Profisafe, provides a hardware-level fallback that halts all motion within
the safety-rated stopping time.

\subsection{AI Model Robustness}
\label{sec:chall_ai}

The accuracy of the defect-detection model depends on the quality of the input
imagery, which proved sensitive to the factory lighting environment.
High-intensity overhead fixtures at Silverline produced specular reflections and
cast shadows on the metallic and glass product surfaces, introducing image noise
that degraded model performance.

Two corrective measures were applied. First, the number of lighting fixtures
within the inspection cell was increased and their positions adjusted to produce
more uniform illumination, reducing both glare and shadow formation. Second, the
training dataset was augmented with images captured under the revised lighting
conditions, improving the model's tolerance to residual reflections. Balancing
illumination levels for human safety (adequate ambient light) against the
requirements of the vision system (controlled, diffuse light with minimal
specular highlights) required iterative adjustment.

After calibration, the hood-detection and tracking module reached an accuracy
approaching 100\% on the validation set, attributable to the high quality of the
curated dataset and the stable camera geometry within the cell. The 2D
defect-detection module reached an accuracy of approximately 85\% against
ground-truth inspections, with the residual error dominated by false positives on
the white-coloured hood variant: backlit button icons on the control panel
exhibited visual signatures similar to certain surface defects against the white
background, causing the detector to flag them spuriously. A polarising anti-glare
lens fitted to the Basler camera suppressed the remaining specular reflections on
the glass surfaces, and the training set was augmented with images captured under
the revised lighting conditions. Further gains would benefit from a larger and
more diverse training set, but this performance was deemed acceptable for the
pilot because every flagged region is visually verified by the operator during
the manual inspection step (see Section~\ref{sec:arch_ai}).

\subsection{Field Integration Challenges}
\label{sec:chall_field}

Several practical difficulties arose during on-site installation and
commissioning, each requiring specific engineering solutions.

\paragraph{Lighting adequacy.}
The existing factory lighting was insufficient for reliable camera acquisition.
Additional LED fixtures were installed on the end effector and within the cell to
ensure adequate and consistent illumination of the inspection zone.

\paragraph{Robot reach limitations.}
Certain surfaces of the product fell outside the UR\,10e's reachable workspace.
The camera platform was repositioned, and the inspection path was re-planned to
cover the required surface area within the robot's kinematic envelope.

\paragraph{Equipment assembly.}
The camera, microphone, lighting, SPS device, and robot had to be assembled and
secured under a tight project timeline. Component procurement delays were
mitigated by pre-staging equipment and establishing a documented assembly
procedure.

\paragraph{Multi-robot namespace management.}
A central technical challenge concerned the coordination of two ROS\,2 robot
stacks. The UR\,10e and the Comau Racer-5 each publish on a standard set of ROS\,2
topic names (joint states, controller commands, tool I/O), and running both
stacks against the same ROS\,2 graph caused namespace collisions that prevented
either robot from operating reliably. Each robot was therefore moved into a
dedicated ROS\,2 namespace, and topic remappings were applied at the orchestrator
level so that subscribers could distinguish UR-side messages from Comau-side
messages unambiguously. Once isolated, the two robots could be brought up
simultaneously, share the same coordination logic, and exchange high-level state
transitions without interference. ROS\,2 message timestamps were retained
throughout the pipeline so that data from the camera, microphone, and SPS device
could be reconstructed against the correct stage of each inspection cycle during
post-hoc analysis.

\paragraph{System optimisation and startup streamlining.}
Initial launches of the cell required the operator to start each ROS\,2 node
manually in the correct order, which was both error-prone and time-consuming. A
hierarchical launch file was therefore designed to orchestrate node bring-up, with
ROS\,2 event handlers used to queue critical nodes so that drivers, AI modules,
the synchronisation layer, and the Grafana dashboard always come up in the correct
order and any process collisions are avoided. To remove the ROS\,2 learning curve
from the daily workflow, a bash launcher script was wrapped behind a single
\textsc{Start}/\textsc{Stop} button embedded in the operator UI, so that the
entire stack can be brought up or torn down by an operator who has never used a
ROS\,2 command line. These changes substantially improved operational reliability
and allow the cell to recover from power cycles or maintenance interventions with
minimal operator intervention.

\section{Experimental Setup and Results}
\label{sec:experiments}

\subsection{Pilot Deployment}
\label{sec:exp_setup}

\begin{figure}[t]
\centering
\begin{minipage}[t]{0.48\textwidth}
\centering
\includegraphics[width=\linewidth,height=5cm,keepaspectratio]{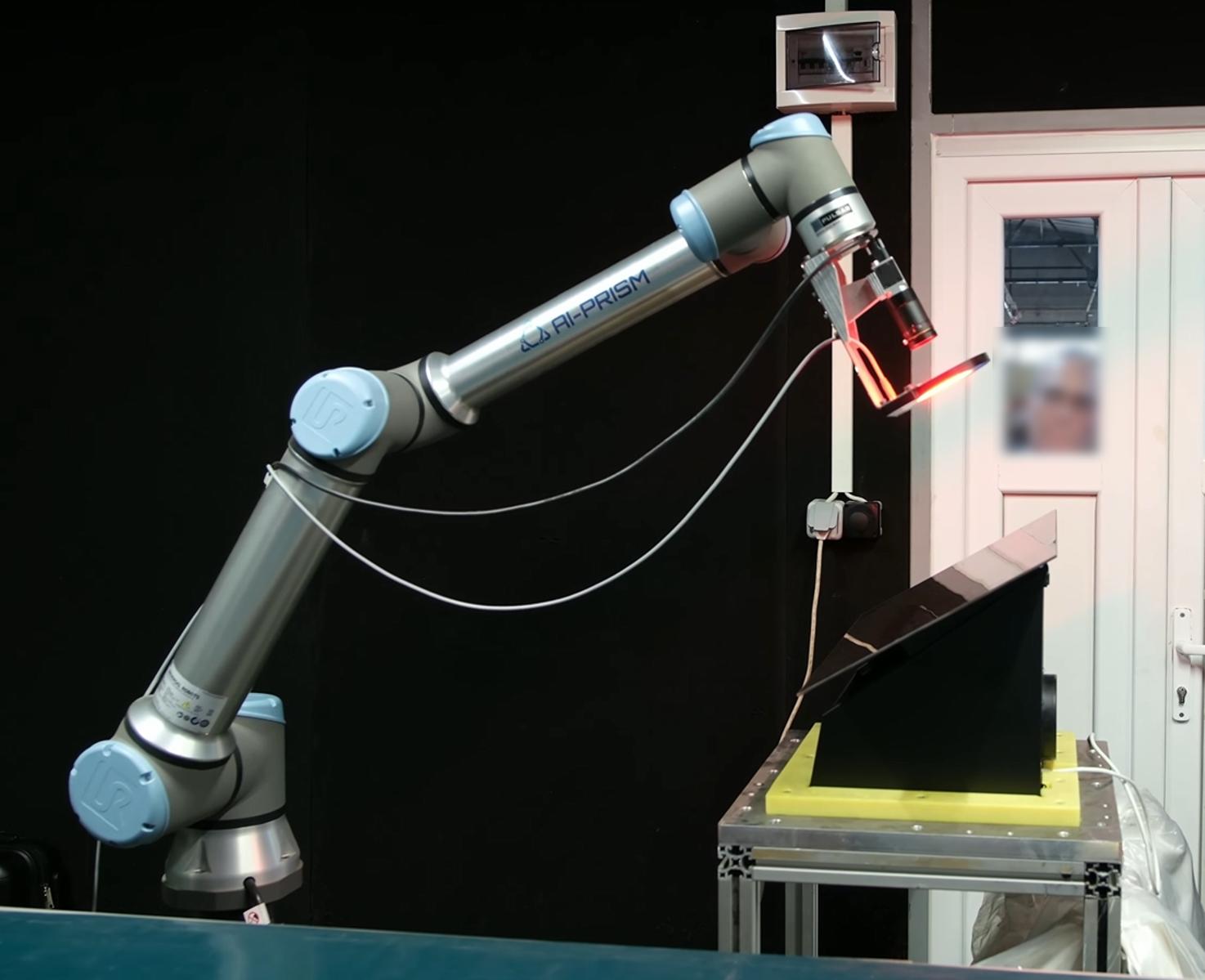}
\caption{Physical deployment of the Silverline inspection cell: the UR\,10e arm
with camera and LED lighting over a sample product panel.}
\label{fig:ur_setup}
\end{minipage}\hfill
\begin{minipage}[t]{0.48\textwidth}
\centering
\includegraphics[width=\linewidth,height=5cm,keepaspectratio]{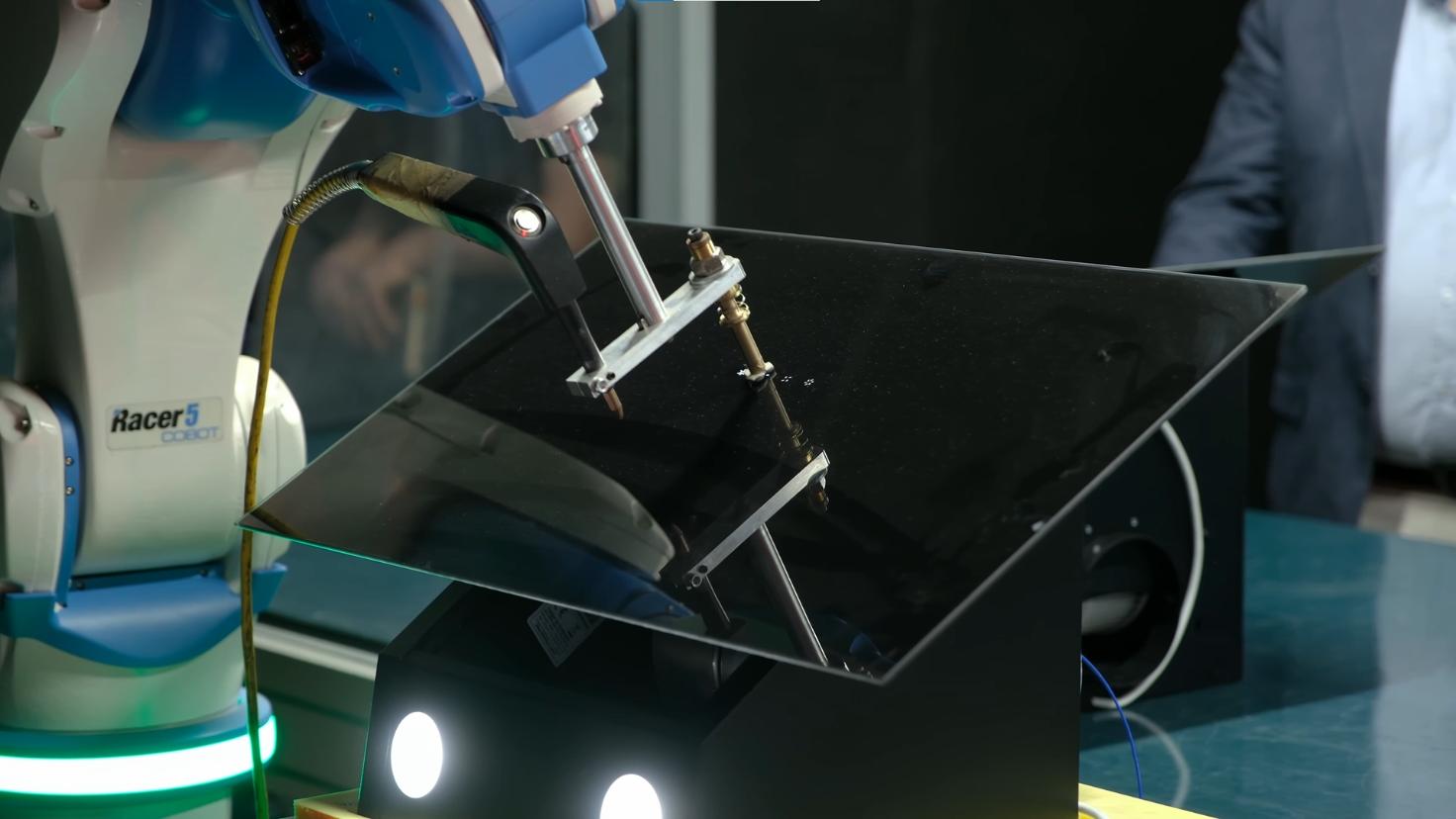}
\caption{The Comau Racer-5 cobot performing functional tests on a
kitchen-appliance hood.}
\label{fig:comau_setup}
\end{minipage}
\end{figure}

The inspection cell was deployed on an active assembly line at the Silverline
factory. Fig.~\ref{fig:ur_setup} shows the physical configuration: the UR\,10e
robot arm is mounted adjacent to the conveyor, with the Basler camera and LED
lighting ring attached to the end effector. The server workstation is positioned
beside the cell, running the ROS\,2 stack, AI inference pipeline, and Grafana
dashboard. A sample kitchen-appliance panel is placed on the inspection platform
for the robot to scan.

The principal hardware components of the cell are summarised in
Table~\ref{tab:hardware}.

\begin{figure}[t]
\centering
\includegraphics[width=0.72\textwidth]{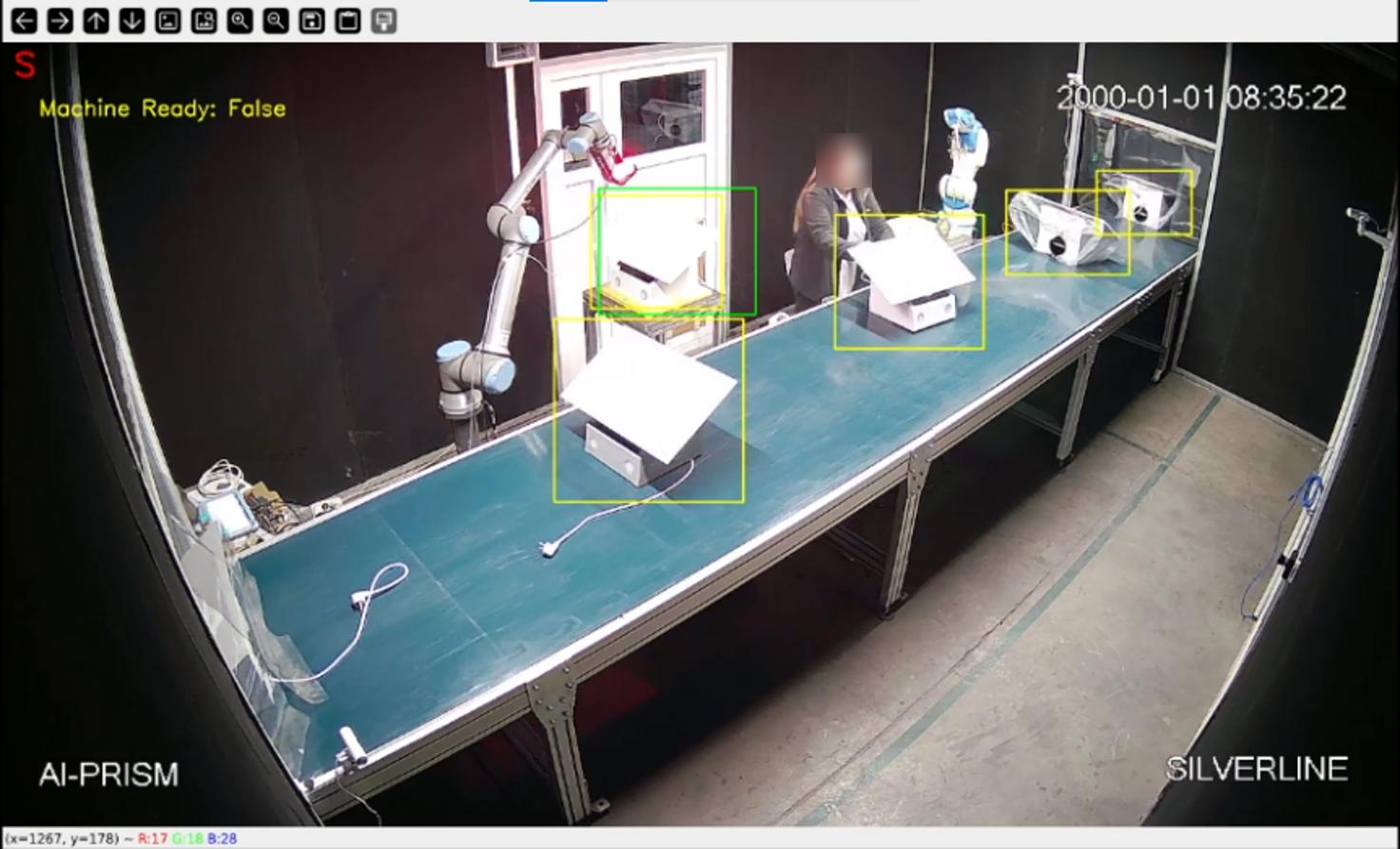}
\caption{Still frame of the tracking AI module showing several white appliances
tracked on the conveyor belt.}
\label{fig:full_setup}
\end{figure}

\begin{table}[t]
\caption{Principal hardware components of the Silverline inspection cell}
\label{tab:hardware}
\centering
\renewcommand{\arraystretch}{1.15}
\begin{tabular}{@{}p{0.30\textwidth}p{0.62\textwidth}@{}}
\toprule
\textbf{Component} & \textbf{Specification} \\
\midrule
Inspection cobot &
Universal Robots UR\,10e collaborative manipulator \\
Functional-test cobot &
Comau Racer-5 collaborative manipulator, TCP/IP driver \\
End-effector camera &
Basler ace industrial camera, USB\,3.0, with anti-glare polarising lens and
annular LED ring \\
Cell-overview camera &
Dahua IP security camera \\
Acoustic sensor &
TIA microphone, interfaced via signal conditioner and a LabJack DAQ \\
Electrical safety tester &
SPS KT1885 electronic safety tester, serial interface \\
Functional safety &
Siemens PLC with HMI panel, Profinet/Profisafe network \\
Server / edge node &
Workstation with NVIDIA RTX\,3060 GPU, running Ubuntu~22.04\,LTS and ROS\,2
Humble, with a local database, InfluxDB, and Grafana \\
\bottomrule
\end{tabular}
\end{table}

\subsection{Evaluation}
\label{sec:exp_results}

The cell was evaluated against the four key performance indicators defined for
Use Case~4 of the AI-PRISM project~\cite{ref_aiprism_ra}: a social KPI on
operator mental workload (U4\_KPI1), a social KPI on performance and job
satisfaction (U4\_KPI2), an economic KPI on resource efficiency (U4\_KPI4), and
an economic KPI on per-unit quality-check time (U4\_KPI5). The results are
summarised in Table~\ref{tab:results}.

\paragraph{Defect-detection performance.}
Validated against ground-truth inspections, the hood-detection and tracking
module achieved an accuracy approaching 100\% on the validation set, attributable
to the high quality of the curated dataset and the stable lighting and camera
geometry within the cell. The 2D defect-detection module achieved an accuracy of
approximately 85\%, with the residual error dominated by false positives on the
white-hood variant. Per-frame inference latency averaged 0.25\,s; the remaining
processing time was attributable to network transfer, yielding an end-to-end
latency of approximately 0.5\,s from frame acquisition to display of detections on
the operator UI, satisfying the real-time requirements of the inspection workflow.

\paragraph{Cycle-time and resource-efficiency analysis.}
The per-unit quality-check time decreased from a baseline of 82\,s to 61\,s with
the AI-PRISM cell in operation, an approximately 25\% reduction that exceeds the
original 10\% target for U4\_KPI5. Resource efficiency at the final-control
station improved from a baseline of 0.75 to 0.88 ($\approx$17\% increase against a
target of $+15$\%), achieved by reducing the number of operators required at the
station from three to one.

\paragraph{Operator feedback and human-factor results.}
Standardised instruments were administered to operators before and after exposure
to the AI-PRISM cell. NASA-TLX scores~\cite{ref_nasa_tlx} showed a statistically
significant reduction in mental demand ($p=0.005$), and reductions in temporal
demand ($p=0.077$) and performance ($p=0.089$) that approached but did not reach
the conventional 0.05 threshold. Eye-tracking measurements recorded an 82\%
reduction in the number of times operators looked at the component during
inspection, indicating substantially less visual workload. Occupational
self-efficacy scores~\cite{ref_self_efficacy} stayed high in both
conditions---5.13 (SD\,$=$\,0.78) in the manual phase and 4.95 (SD\,$=$\,0.68) with
the AI-PRISM cell---indicating that participants continued to feel capable and
prepared in their job regardless of the level of automation. Technology readiness
scores~\cite{ref_tri}, by contrast, were moderate in both settings but lower for
the AI-PRISM condition (2.88, SD\,$=$\,0.38) than for the manual condition (3.18,
SD\,$=$\,0.59). Qualitative interview data echoed this asymmetry: while operators
acknowledged practical advantages such as reduced noise and physical effort,
several reported feeling less directly engaged when the cobots took over part of
the task and expressed caution about delegating quality decisions to the AI. The
cognitive-ergonomics target (U4\_KPI1) was therefore clearly met, while broader
technology acceptance remains an area for continued attention as operators gain
familiarity with the system.

\paragraph{System reliability.}
A live demonstration of the cell was conducted on October~8, 2025, with attendees
from academia, industry, and the AI-PRISM consortium. The cell ran the full
automated inspection and functional-test sequence on representative hoods without
manual intervention, including hood tracking and counting, defect detection on the
glass surface, grounding tests, and motor-speed and throttling tests with acoustic
monitoring. Recorded measurements were displayed in real time on the
Grafana-based operator dashboard and persisted in the local database for
traceability.

\begin{table}[t]
\caption{Key performance indicators of the Silverline inspection cell}
\label{tab:results}
\centering
\renewcommand{\arraystretch}{1.15}
\begin{tabular}{@{}p{0.46\textwidth}p{0.20\textwidth}p{0.26\textwidth}@{}}
\toprule
\textbf{Indicator} & \textbf{Baseline} & \textbf{Achieved} \\
\midrule
Hood detection \& tracking accuracy & --- & $\approx$\,100\% \\
2D defect-detection accuracy & --- & $\approx$\,85\% \\
Per-frame inference latency & --- & $\approx$\,0.25\,s \\
End-to-end perception latency & --- & $\approx$\,0.5\,s \\
Per-unit QC time (U4\_KPI5) & 82\,s & 61\,s ($-25$\%) \\
Resource efficiency (U4\_KPI4) & 0.75 (3 ops) & 0.88 (1 op, $+17$\%) \\
NASA-TLX mental demand (U4\_KPI1) & baseline & sig.\ reduction ($p=0.005$) \\
Operator viewing time on component & baseline & $-82$\% \\
Occupational self-efficacy (U4\_KPI2) & 5.13 (manual) & 4.95 (AI-PRISM) \\
Technology readiness & 3.18 (manual) & 2.88 (AI-PRISM) \\
\bottomrule
\end{tabular}
\end{table}

\section{Conclusion and Future Work}
\label{sec:conclusion}

This paper presented the design, integration, and field deployment of an
AI-assisted collaborative inspection cell at the Silverline kitchen-appliance
factory, combining a UR\,10e cobot, a machine-vision defect-detection pipeline,
and multi-modal data acquisition under a ROS\,2-based control architecture. A
structured account of the deployment challenges---spanning safety, AI robustness,
lighting, driver compatibility, and dependency management---was provided together
with the engineering solutions applied in each case, and the four-level HRI
analysis showed how ROS\,2 can serve as a unifying framework from task-level
programming through process-level production coordination.

The pilot confirmed that retrofitting an existing assembly line with AI-augmented
cobots is technically feasible but requires attention to integration issues that
are seldom reported in algorithm-centric publications. The human-centered design
of the cell, which augments rather than replaces the operator, proved important
both for workforce acceptance and for maintaining flexibility in production.

Quantitatively, the deployed cell achieved hood-tracking accuracy approaching
100\%, defect-detection accuracy of approximately 85\%, a 25\% reduction in
per-unit quality-check time, and a 17\% improvement in resource efficiency,
alongside a statistically significant reduction in operator mental demand and an
82\% reduction in operator visual-inspection viewing time. The principal
trade-off observed was a modest sensitivity of detection accuracy to product
variant under factory-floor lighting, and a small but consistent drop in operator
technology-readiness scores after exposure to the cell, indicating that the
economic case is tied to both the production-volume mix and continued investment
in dataset curation and operator acceptance.

Several directions for future work follow. First, an online learning mechanism
could let the defect-detection model adapt incrementally to new product variants
and evolving surface finishes without full retraining. Second, the sensor
synchronization framework should be extended to support tighter temporal
alignment across modalities, enabling fused audio-visual anomaly detection.
Third, scaling the architecture to multiple inspection cells on parallel lines
will test the system's modularity and the network infrastructure's capacity.

\begin{credits}
\subsubsection{\ackname}
This work was supported by the AI-PRISM project (AI-Powered human-centred Robot
Interactions for Smart Manufacturing), which received funding from the European
Union's Horizon Europe research and innovation programme under grant agreement
No.~101058589. The authors thank Silverline End\"ustri ve Ticaret A.\c{S}. for
providing access to the production facility and supporting the pilot deployment,
and the AI-PRISM consortium partners for their contributions to the underlying
technical components. The authors also thank Dr Iveta Eimontaite from Cranfield University for her contributions to the analysis of the social impact of the AI-PRISM solution. 

\subsubsection{\discintname}
The authors have no competing interests to declare that are relevant to the
content of this article.
\end{credits}

\end{document}